\documentclass[runningheads]{llncs}

\usepackage[T1]{fontenc}
\usepackage{amssymb}
\usepackage{amsmath}
\usepackage{amsfonts}
\usepackage{graphicx}
\usepackage{url}
\usepackage{multirow}
\usepackage{booktabs}

\DeclareMathOperator*{\argmin}{arg\,min}

\def\doi#1{\href{https://doi.org/\detokenize{#1}}{\url{https://doi.org/\detokenize{#1}}}}
\usepackage{graphicx}
\usepackage{listings}
\begin{document}

\title{An improved neural network model for treatment effect estimation}

\titlerunning{An improved neural network model for treatment effect estimation}

\author{Niki Kiriakidou\thanks{Corresponding author}\orcidID{0000-0003-1729-4124} \and
		Christos Diou\orcidID{0000-0002-2461-1928}}
	\authorrunning{N. Kiriakidou and C. Diou}

\institute{Department of Informatics and Telematics, Harokopio University, Greece 
\email{\{kiriakidou,cdiou\}@hua.gr}}

\maketitle              
\begin{abstract}
Nowadays, in many scientific and industrial fields there is an increasing need for estimating treatment effects and answering causal questions. The key for addressing these problems is the wealth of observational data and the processes for leveraging this data. In this work, we propose a new model for predicting the potential outcomes and the propensity score, which is based on a neural network architecture. The proposed model exploits the covariates as well as the outcomes of neighboring instances in training data. Numerical experiments illustrate that the proposed model reports better treatment effect estimation performance compared to state-of-the-art models.
\end{abstract}

\section{Introduction}

For decades, causal inference has been a crucial research topic in many scientific fields, such as healthcare \cite{glass2013causal}, education \cite{gustafsson2013causal} and economics \cite{varian2016causal}. Causal inference aims at answering questions regarding the effect of interventions, (e.g., a new drug, a new educational method or a new pricing policy) to the target outcome variables (e.g., health, learning or financial indicators, respectively). 

The inference of causal effects is a challenging problem and the most effectual way to infer causality is through randomized controlled trials (RCTs). In many cases, however, it is expensive, time-consuming, unethical or even impossible to conduct an RCT.  Nowadays, the abundance of observational data presents an opportunity for accurate estimation of causal effects, however, observational data contain recorded information about samples, such as actions and outcomes along with appropriate context, but there is way to directly influence the mechanism that caused the action. Furthermore, in observational data may exist confounding variables, which affect both treatment and outcome. If these are not adjusted, they could lead to incorrect and misleading results.



%

In this work, a neural network model is proposed for treatment effect estimation through the prediction of the conditional outcomes and the propensity score. The model extends the state-of-the-art Dragonnet architecture \cite{shi2019adapting} to exploit the covariates along with information from the outcomes of the instances contained in the training data.
The rationale behind the proposed approach is to enrich the inputs of the model with the average outcomes of the nearest neighbors from the control and treatment group along with the covariates, in order to reduce bias and increase the prediction accuracy.
To estimate treatment effects, the proposed method first trains a model for the prediction of conditional outcomes and the propensity score and then the trained model is used by a downstream estimator. Our experiments illustrate that the proposed approach maintains state of the art performance for the estimation of average treatment effect (ATE), while it leads to significant improvement in estimating the individual treatment effect (ITE).

The remainder of this paper is organised as follows: Section \ref{Sec:2} presents a review of neural network based models for the estimation of treatment effects. Section \ref{Sec:3} presents a comprehensive description of the proposed modified model and its architecture. Section \ref{Sec:4} provides information about the data. Section \ref{Sec:5} presents a detailed experimental analysis, focusing on the evaluation of the proposed model. 
Section \ref{Sec:6} summarizes the main findings and conclusions of this research, and some interesting directions for future work. 

\section{Related Work}\label{Sec:2}

During the last decade, a lot of research has been conducted towards more accurate and reliable estimation of treatment effects. Most of this research is based on the use of neural networks, exploiting the predictive power of these machine learning models. 

Johansson et al. \cite{johansson2016learning} proposed a new algorithmic framework for counterfactual inference. More specifically, they formulated the causal inference problem as a domain adaptation problem and developed a new class of representation algorithms for the calculation of treatment effects. They highlighted that learning representations, which enforce similarity between control and treated groups, is able to lead to better estimations of causal effects. They compared a variant of the proposed algorithm based on a neural network approach, named Balancing Neural Network (BNN), against traditional models, which reported the best overall performance.

Shalit et al. \cite{shalit2017estimating} proposed a new theoretical analysis and a new framework, named Counterfactual Regression (CFR) for predicting individual treatment effects. The proposed framework aims on learning a balanced representation using a prediction model, so that the distributions of control and treated group look similar. To measure the distances between two distributions they utilized the integral probability metrics: Maximum Mean Discrepancy (MMD) \cite{gretton2012kernel} and Wasserstein distance (Wass) \cite{villani2009optimal}. Additionaly, the major contribution of their work is the introduction of a generalization-bound for the estimation of individual treatment effect, where every individual is only identified by its features. In their experiments, they compared the performance of two proposed models, CFR (MMD) and CFR (Wass), which use MMD and Wass distances, respectively, against state-of-the-art models. Furthermore, they included 
a variant without balance regularization, named Treatment Agnostic Representation Network (TARNet). Based on their experimental analysis, they stated that all proposed models presented the best performance in terms of estimating treatment effects. 

Another approach for estimating individual treatment effect was proposed by Yoon et al. \cite{yoon2018ganite}, which is based on Generative Adversarial Nets (GANs). The rationale behind the proposed approach is to simulate the uncertainty in the counterfactual distributions by considering learning them using a GAN model. Along this line, they developed a novel model, named Generative Adversarial Nets for inference of Individalized Treatment Effects (GANITE), which was able to provide confidence intervals for its predictions. Their numerical experiments revealed that the proposed method exhibited promising performance.

Louizos et al. \cite{louizos2017causal} highlighted the significance of handling confounders for inferring treatment effects from observational data. More specifically, they stated that there is a strong possibility of existing uncertain and noisy ``proxy variables'', in case there is no access to all confounders. To address the previous difficulties they proposed a new model, called CEVAE, based on variational autoencoders. A considerable advantage of their approach is that the data generating process as well as the structure of the hidden confounders requires substantially weaker assumptions. Finally, the authors presented that CEVAE exhibited more robust behaviour against hidden confounders in the case of noisy proxies.

Shi et al. \cite{shi2019adapting} proposed a novel neural network model for estimating treatment effects from observational data. The proposed model, named Dragonnet, focuses on improving the estimations through the sufficiency of the propensity score. Additionally, the authors proposed targeted regularization, which constitutes a procedure to induce bias based on non-parametric estimation theory and aims to further improve the estimation of treatment effect. Finally, the authors provided experimental evidence about the superiority of Dragonnet against BNN, CEVAE, GANITE, TARNet, CFR (MMD) and CFR (Wass) using two benchmark datasets.

In this work, we propose a neural network model for predicting the potential outcomes and the propensity score. The proposed model architecture is a modification of Dragonnet's architecture. The major difference between the proposed model and Dragonnet is that the former's inputs contain information from the covariates as well as from the outputs of control and treated group.Our numerical experiments provide empirical and statistical evidence about the efficacy and efficiency of our approach.   


\section{Modified Dragonnet model}\label{Sec:3}

In this section, we present the proposed model for the estimation of treatment effects. The rationale behind our approach is to enrich the training data with information from the outcomes, which can be exploited by the proposed model in order to obtain more accurate predictions.

\subsection{Calculation of average outcome vectors}

We limit our discussion to the case of binary treatments. Let $\mathcal{X}$ denote the $d$-dimensional space of covariates and  consider a joint distribution $\Pi$ on $\mathcal{X}\times \{ 0, 1\} \times \mathcal{Y}$.
Suppose that $(X, T, Y) \sim \Pi$, are random variables with domains $\mathcal{X}$, $\{0, 1\}$ and $\mathcal{Y}$, corresponding to the covariates, treatment and outcome for a single sample, respectively. 
Let also $Y_0$ denote the outcome for a sample when $T=0$ and $Y_1$ stand for the outcome of a sample when $T=1$.

Given a dataset $(x_i,t_i,y_i)$, $i=1,2,\dots,n$ where $x_i \in \mathcal{X}$, $t_i \in \{0,1\}$ and $ y_{i} \in \mathcal{Y}$ our goal is to estimate the average treatment effect 

\begin{equation}
         \psi = E[Y\,|\, X, T=1]- E[Y\,|\, X, T=0]
\end{equation}

For each observed sample in the dataset, either $t_i=0$ ($Y_0$ is factual) or $t_i=1$ ($Y_0$ is counterfactual) and  $y_i = t_i Y_1 + (1-t_i)Y_0$, based on the framework of Neyman-Rubin \cite{rubin2005causal}.


%
The main idea of our model is to reduce bias in treatment effect estimation, by utilizing the average outcomes of $k$ nearest neighbors $\overline{y}_{i}^{(0)}$ of the control group and $\overline{y}_{i}^{(1)}$ of the treatment group for each available sample $i$.

Algorithm 1 presents a pseudocode for the calculation of  $\overline{y}_{i}^{(0)}$ and $\overline{y}_{i}^{(1)}$. The algorithm takes as inputs the design matrix $\mathbf{X}$, whose rows correspond to the covariate vectors of samples, the binary vector of treatment values $\mathbf{t}$, the outcome vector $\mathbf{y}$ in the dataset, as well as the number of nearest neighbors $k$.

Initially, $\overline{\mathbf{y}}^{(0)}$ and $\overline{\mathbf{y}}^{(1)}$ are initialized to $\mathbf{0}$. (Step 1). Next, for every instance $\mathbf{x}_{i}$ we calculate the average outcomes for control and treated group (Steps 2-7). More specifically, we calculate the k-nearest neighbors of $\mathbf{x}_{i}$ in $\mathbf{X}$, contained in the control group (i.e $T=0$) and append their corresponding indices in the index set $S_0$ (Step 4). Then, we calculate the average of the outcomes of these neighbors, $\overline{y}_i^{(0)} = \frac{1}{k} \sum_{j\in \mathcal{S}_0} y_j$  (Step 5) 
%
Similarly, we calculate the average outcome of the $k$-nearest neighbors of $\mathbf{x_{i}}$, contained in the treatment group (i.e $T=1$)(Step 6-7)\\

%
%


\noindent{}\textbf{Algorithm 1}

\noindent\rule{12cm}{0.4pt}
\begin{description}
	\item[Inputs:]
	\item \quad $\mathbf{X}$: design matrix      
	\item \quad $\mathbf{t}$: vector of treatment values $t$
	\item \quad $\mathbf{y}$: vector of outcome values $y$
	\item \quad $k$: number of nearest neighbors\vspace{.2cm}
	
	\item[Output:]
	\item \quad $\overline{\mathbf{y}}^{(0)}$: vector with average of $k$-nearest outcomes from control group for each sample
	\item \quad $\overline{\mathbf{y}}^{(1)}$: vector with average of $k$-nearest outcomes from treatment group for each sample
	
	\setlength\itemsep{.3em}
	\vspace{2cm}
	\item[Step 1:] Set  $\overline{\mathbf{y}}^{(0)} = \mathbf{0}$ and $\overline{\mathbf{y}}^{(1)} =\mathbf{0}$
	\item[Step 2:] for $\mathbf{x}_{i}, i=1,2,\dots,n$ do 
	\item[Step 3:] \hspace{.4cm} $\mathbf{x}_{i}=\mathbf{X}[i,:]$
	\item[Step 4:] \hspace{.4cm} Calculate the index set $\mathcal{S}_0$ containing the indices of the 
	\item[] \hspace{1.65cm} $k$-nearest neighbors of $\mathbf{x}_{i}$ with $T = 0$
	\item[Step 5:] \hspace{.4cm} $\displaystyle \overline{y}_i^{(0)} = \frac{1}{k} \sum_{j\in \mathcal{S}_0} y_j $
	%
	
	%
	\item[Step 6:] \hspace{.4cm} Calculate the index set $\mathcal{S}_1$ containing the indices of the
	\item[] \hspace{1.65cm} $k$-nearest neighbors of $\mathbf{x}_{i}$  with $T = 1$
	\item[Step 7:] \hspace{.4cm} $\displaystyle\overline{y}_i^{(1)} = \frac{1}{k} \sum_{j\in \mathcal{S}_1} y_j $
	%
	
\end{description}
\noindent\rule{12cm}{0.4pt}\\

Based on the presented iterative process the average outcome from control and treated group is obtained and stored in $\overline{\mathbf{y}}^{(0)}$ and $\overline{\mathbf{y}}^{(1)}$, respectively. Notice that these will be used by the proposed model for the prediction of the conditional outcomes $Q(t,\mathbf{x}) = E(Y\, |\, X =\mathbf{x}, T=t)$ and the propensity score $g(x) = P(T = 1\, |\, X =\mathbf{x})$.

\subsection{Modified Dragonnet architecture}

The proposed model consists of a modification of the state-of-the-art Dragonnet model \cite{shi2019adapting}. The model takes as inputs the design matrix $\mathbf{X}$ and the average outcomes from control and treated group, $\overline{\mathbf{y}}^{(0)}$ and $\overline{\mathbf{y}}^{(1)}$, respectively, while its three-headed architecture produces the predictions of  propensity score $\hat{g}(\cdot)$ and conditional outcomes $\hat{Q}(0, \cdot, \cdot, \cdot)$ and $\hat{Q}(1, \cdot, \cdot, \cdot)$.

\begin{figure}[!ht]
	\centering
	\includegraphics[width=0.8\linewidth]{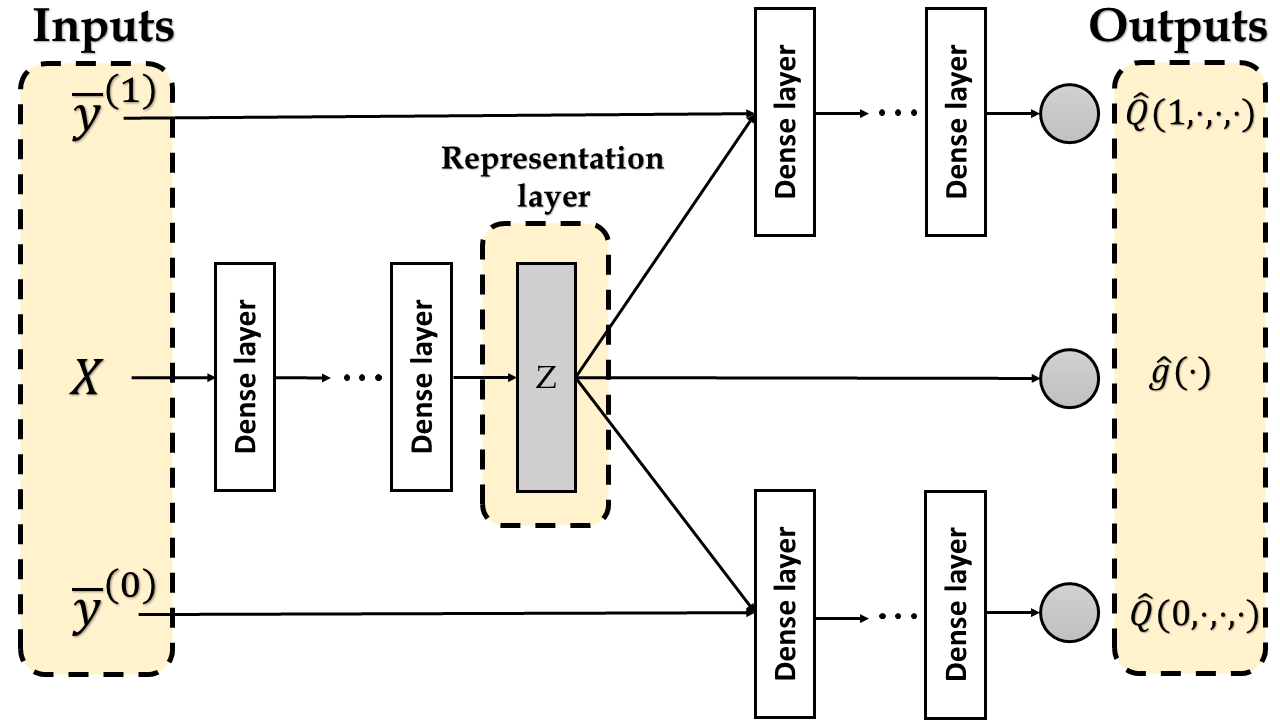}
	\caption{Modified Dragonnet architecture}\label{Fig:Modified Dragonnet}
\end{figure}

Figure \ref{Fig:Modified Dragonnet} presents a high-level architecture of the proposed modified Dragonnet model.
Initially,  a number of dense layers are utilized in order to produce a representation layer $Z(\mathbf{X}) \in \mathbb{R}^p$. Next, the output of $Z(\mathbf{X})$ is concatenated with $\overline{\mathbf{y}}^{(0)}$ and the combined information is further processed by dense layers for the prediction of the outcome $\hat{Q}(0, \cdot, \cdot, \cdot)$. Similarly, the output of $Z(\mathbf{X})$ is concatenated with $\overline{\mathbf{y}}^{(1)}$ and through  a number of dense layers the model provides the outcome $\hat{Q}(1, \cdot, \cdot, \cdot, \cdot)$. Additionally, the shared representation $Z(\mathbf{X})$ is used for predicting $\hat{g}(\cdot)$, through the use of a simple linear map followed by a sigmoid activation function.

The model is trained by minimizing the following loss function
\begin{equation}\label{Eq:Loss}
       \hat{\theta} =	\argmin_\theta \hat{R}(\theta; \mathbf{X}, \overline{\mathbf{y}}^{(0)}, \overline{\mathbf{y}}^{(1)})
\end{equation}
where $\theta$ is the parameter vector and $\hat{R}$ is defined by
\begin{equation}\label{Eq:R}
	\hat{R}(\theta; \mathbf{X}, \overline{\mathbf{y}}^{(0)}, \overline{\mathbf{y}}^{(1)}) = \frac{1}{n} \sum_i \left[ ( Q^{nn}(t_i,\mathbf{x}_{i},\overline{y}_i^{(0)},\overline{y}_i^{(1)};\theta) -y_i )^2 + \alpha f(g^{nn}(\mathbf{x}_{i};\theta), t_i)\right]
\end{equation}
where $Q^{nn}(t_i, \mathbf{x}_{i}, \overline{y}_i^{(0)}, \overline{y}_i^{(1)};\theta)$ and $g^{nn}(\mathbf{x}_{i}; \theta)$ are the output heads, $f$ is the cross entropy function and $\alpha > 0 $ is a hyperparameter used for weighting the two loss components.\vspace{-.5cm}

Additionally, in order to increase the performance of the proposed model
we utilized \textit{targeted regularization} \cite{shi2019adapting}, which constitutes a modification to the loss function \eqref{Eq:Loss}, by introducing a regularization term and an extra parameter.

More specifically, the modified Dragonnet model is trained by minimizing the following loss
\begin{equation}\label{Eq:Regularized loss}
	\hat{\theta},\hat{\epsilon} = 
		\argmin_{\theta,\epsilon} \left[ 
		\hat{R}(\theta; \mathbf{X}, \overline{\mathbf{y}}^{(0)}, \overline{\mathbf{y}}^{(1)}) + \beta\frac{1}{n} \sum_i
		\gamma(y_i, t_i, \mathbf{x}_{i}, \overline{y}_i^{(0)}, \overline{y}_i^{(1)}; \theta,\epsilon)
		\right]
\end{equation}
where $\beta,\epsilon$ are positive parameters,
$\hat{R}(\theta; \mathbf{X}, \overline{\mathbf{y}}^{(0)}, \overline{\mathbf{y}}^{(1)})$ is defined by Eq.~\eqref{Eq:R} and the regularization term
$\gamma(y_i, t_i, \mathbf{x}_{i}, \overline{y}_i^{(0)}, \overline{y}_i^{(1)}; \theta,\epsilon)$ is defined by
\begin{eqnarray*}
	\gamma(y_i, t_i,\mathbf{x}_{i}, \overline{y}_i^{(0)}, \overline{y}_i^{(1)}; \theta,\epsilon) & = & (y_i - \tilde{Q}(t_i, \mathbf{x}_{i}, \overline{y}_i^{(0)}, \overline{y}_i^{(1)};\theta,\epsilon))^2  \\
	\tilde{Q}(t_i,\mathbf{x}_{i}, \overline{y}_i^{(0)}, \overline{y}_i^{(1)};\theta,\epsilon) &=& Q^{nn}(t_i, \mathbf{x}_{i}, \overline{y}_i^{(0)}, \overline{y}_i^{(1)};\theta) 
	+ \epsilon\left[ \frac{t_i}{g^{nn}(\mathbf{x}_{i};\theta)} -
	\frac{1-t_i}{1-g^{nn}(\mathbf{x}_{i};\theta)} \right]
\end{eqnarray*}
The rationale behind the loss function \eqref{Eq:Regularized loss} is based on non-parametric estimation theory and consists on improving the model's estimation of treatment effects. Additionally, under conditions, the following estimator of $\psi$ 
$$ 
	\hat{\psi}^\text{treg} = \frac{1}{n} \sum_{i=1}^{n}[ \hat{Q}^{treg}(1,\mathbf{x}_{i},\overline{y}_i^{(0)}, \overline{y}_i^{(1)})-\hat{Q}^{treg}(0,\mathbf{x}_{i},\overline{y}_i^{(0)}, \overline{y}_i^{(1)})  ]
$$ 
where $\hat{Q}^{treg} = \tilde{Q}(\cdot,\cdot,\cdot,\cdot;\hat{\theta},\hat{\epsilon})$, 
has the following properties \cite{van2011targeted} :
\begin{enumerate}
	\item  $\hat\psi$ will fast converge to $\psi$ even in case $\hat{Q}$ and $\hat{g}$ converge slowly to $Q$ and $g$.
	\item asymptotically  $\hat\psi$ has the lowest variance from any other considered estimator of $\psi$.
\end{enumerate}

\section{Data}\label{Sec:4}

Considering that real-world data for causal inference are rarely available, we scarcely have access to the ground truth causal effects. Therefore, to overcome this problem we rely on semi-synthetic data for the empirical evaluation of causal estimation procedures.

We used the semi-synthetic IHDP dataset introduced by Hill \cite{hill2011bayesian}.
This dataset was constructed from the Infant Health and Development Program and 
the outcome and treatment assignment are fully known. It comprises 25 features
regarding childs and mothers and 747 units, in which 139 belong to the treatment group and the rest 608 belong to the control group. 
In order to have comparable results, we used 1000 realizations from the NPCI package \cite{dorie2016npci} similar to Shi et al. \cite{shi2019adapting}.

\section{Experimental results}\label{Sec:5}

In this section, we evaluate the prediction performance of the proposed modified Dragonnet model against the state-of-the-art Dragonnet model. It is worth mentioning, that we selected to compare the proposed model against Dragonnet, since it outperforms all other state-of-the-art models.

The performance of each model was measured using the metrics absolute error in ATE \cite{shalit2017estimating} $|\epsilon_{ATE}|$ and expected Precision in Estimation of Heterogeneous
Effect \cite{hill2011bayesian}  $\epsilon_{PEHE}$, which are respectively defined by:
$$
|\epsilon_{ATE}| = \left| \frac{1}{n}\sum_{i=1}^{n}[Q(1,\mathbf{x}_{i})-Q(0,\mathbf{x}_{i})]- \hat{\psi}^\text{treg} \right|
%
$$
and 
$$
\epsilon_{PEHE} = \frac{1}{n} \sum_{i=1}^{n} \left[ (Q(1,\mathbf{x}_{i})-Q(0,\mathbf{x}_{i})) - (\hat{Q}^\text{treg}(1,\mathbf{x}_{i},\overline{y}_i^{(0)},\overline{y}_i^{(1)}) -\hat{Q}^\text{treg}(0,\mathbf{x}_{i},\overline{y}_i^{(0)},\overline{y}_i^{(1)}))\right]^{2} 
$$
It is worth noticing, that $|\epsilon_{ATE}|$ and $\epsilon_{PEHE}$ metrics are used to compare the evaluated models as estimators and predictors, respectively and have been also used in \cite{johansson2016learning,louizos2017causal,shalit2017estimating,yoon2018ganite}.

In our experiments, the state-of-the art model Dragonnet was
used with its default optimized parameter settings \cite{shi2019adapting}, while the proposed model followed a similar architecture and hyper-parameter selection with Dragonnet. More specifically, we utilize three dense layers (of 200 neurons with Exponential Linear Unit (ELU) activation function) in order to produce a representation layer $Z(\mathbf{X})$. Next, the output of $Z(\mathbf{X})$ is concatenated with $\overline{\mathbf{y}}^{(0)}$ and the combined information is further processed by two dense layers (of 100 neurons each with ELU activation function and kernel regularizer of $10^{-2}$) for the prediction of the outcome of the control group. A similar approach was used for providing the outcome of the treated group. The hyperperameters were set
as $k = 10$, $\alpha = 1$ and $\beta = 1$ and 
20\% of the training data were utilized for validation as in Dragonnet.
Both evaluated models were trained using stochastic gradient descent with momentum \cite{qian1999momentum}.

The performance of the proposed modified Dragonnet utilizing three different distance metrics i.e Euclidean, Manhattan and Chebychev. These distances constitute the most widely used in the literature \cite{pandit2011comparative,singh2013k}. 
It is worth mentioning that these distances belong to the class of Minkowski distances, which is defined by
$$
	\| x-y\|_p = \left( \sum_i^d {|x_i-y_i|^p} \right)^\frac{1}{p}
$$
where $x,y\in\mathbb{R}^d$ and $p\in\mathbb{N}^*$. In case, $p=1$, $p=2$ and $p=\infty$ the Minkowski distance is reducted to the Manhattan, Euclidean and Chebychev metric, respectively. 
The detailed experimental results for each model and realization of IHDP can be found in \url{https://github.com/kiriakidou/Modified_Dragonnet}.

The implementation code was written in Python 3.7 using Keras library 
\cite{gulli2017deep} and run on a PC (3.2GHz Quad-Core processor, 16GB RAM)
using Windows operating system.

Given into consideration that a small number of simulations tend to dominate
the benchmarking process, the cumulative total for a performance metric over all simulations seem to be too uninformative and misleading.
For this reason, we used Dolan and Moré's \cite{dolan2002benchmarking} performance profiles, which removes the influence of such simulations on the benchmarking process
and provides us information such as probability of success, efficiency and robustness in compact form. In more detail, each profile plots the fraction $P$ of simulations for which any given model is within a factor $\tau$ of the best model.
Additionally, in order to examine and reject the hypothesis that both models perform equally and provide statistical evidence about the superiority of the proposed model,
we utilize the methodology presented in \cite{livieris2019employing}. More specifically, we apply the non-parametric Friedman Aligned-Ranks (FAR) test \cite{hodges2012rank} in order to rank the models and the post-hoc Finner test \cite{finner1993monotonicity} for examining the existence of significant differences.  
Next, we evaluate the performance of: 
\begin{itemize}
	\item ``\textsf{Dragonnet}'', which stands for Dragonnet model of Shalit et al. \cite{shi2019adapting}.\vspace{.2cm}
	
	\item ``\textsf{Modified Dragonnet (Euclidean)}'', which stands for the proposed model using Euclidean distance for the calculation of the average of the outcomes of nearest instances.\vspace{.2cm}
	
	\item ``\textsf{Modified Dragonnet (Manhattan)}'', which stands for the proposed model using Manhattan distance for the calculation of   the average of the outcomes of nearest instances.\vspace{.2cm}
	
	\item ``\textsf{Modified Dragonnet (Chebychev)}'', which stands for the proposed model using Chebychev distance for the calculation of   the average of the outcomes of nearest instances.  
	
\end{itemize}

\begin{figure}[!ht]
	\centering
		\includegraphics[width=0.8\linewidth]{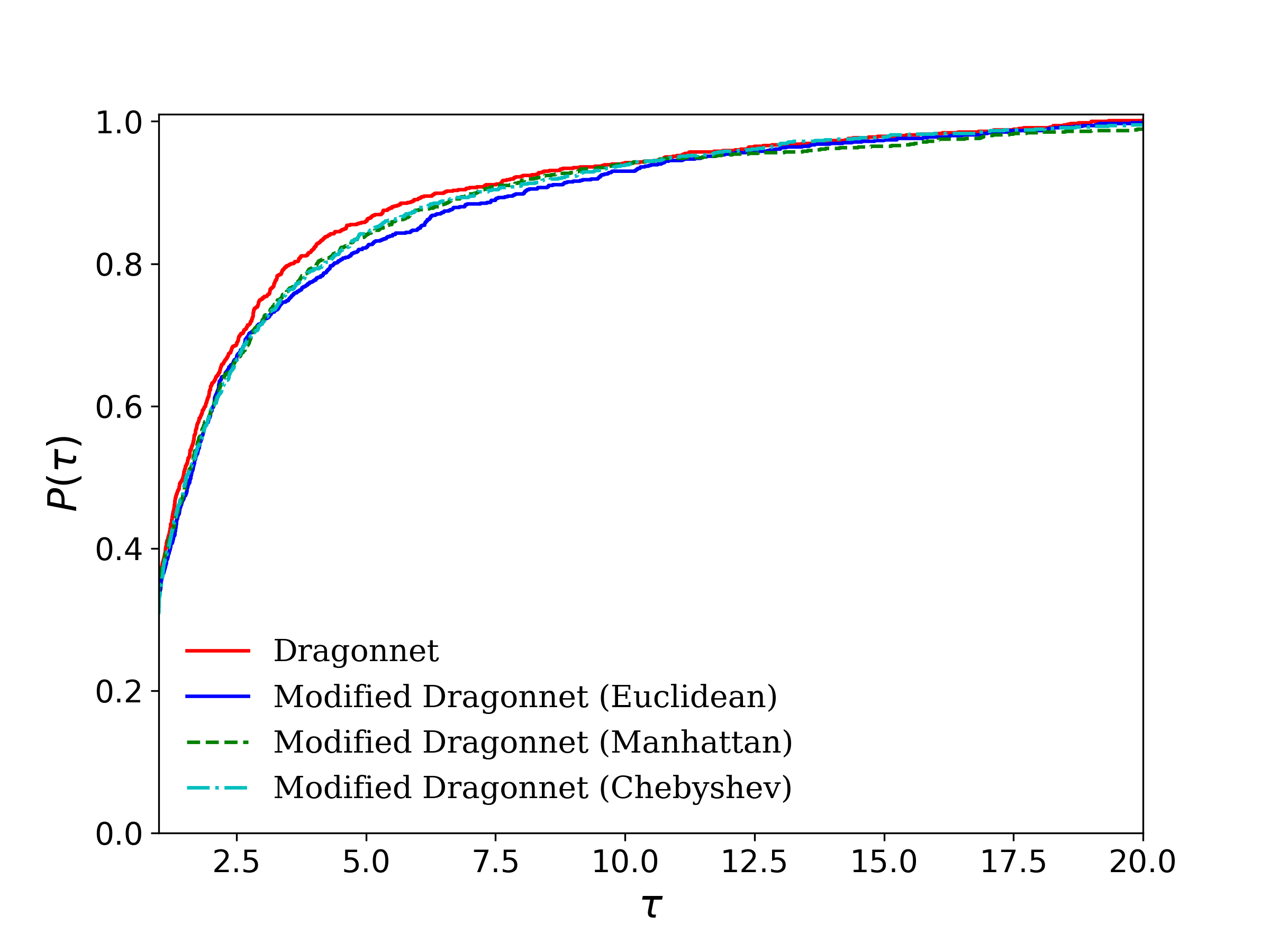}
		\vspace{-.5cm}
	\caption{Performance profiles of all evaluated models based on $|\epsilon_{ATE}|$}\label{Fig:Performance profile ATE}
\end{figure}

Figure \ref{Fig:Performance profile ATE} presents the performance profiles of the three versions of the proposed model and the \textsf{Dragonnet}, based on $|\epsilon_{ATE}|$ metric. Obviously, all compared models reported similar performance. More specifically,  \textsf{Modified Dragonnet (Euclidean)} solves 30\% of the simulations with the lowest error ATE, while both \textsf{Dragonnet} and \textsf{Modified Dragonnet (Chebychev)} solve 28\%. Additionally, \textsf{Modified Dragonnet (Manhattan)} reported the worst performance solving 25\% of simulations. \vspace{-1.2cm}

\begin{figure}[!ht]
	\centering
	\includegraphics[width=0.8\linewidth]{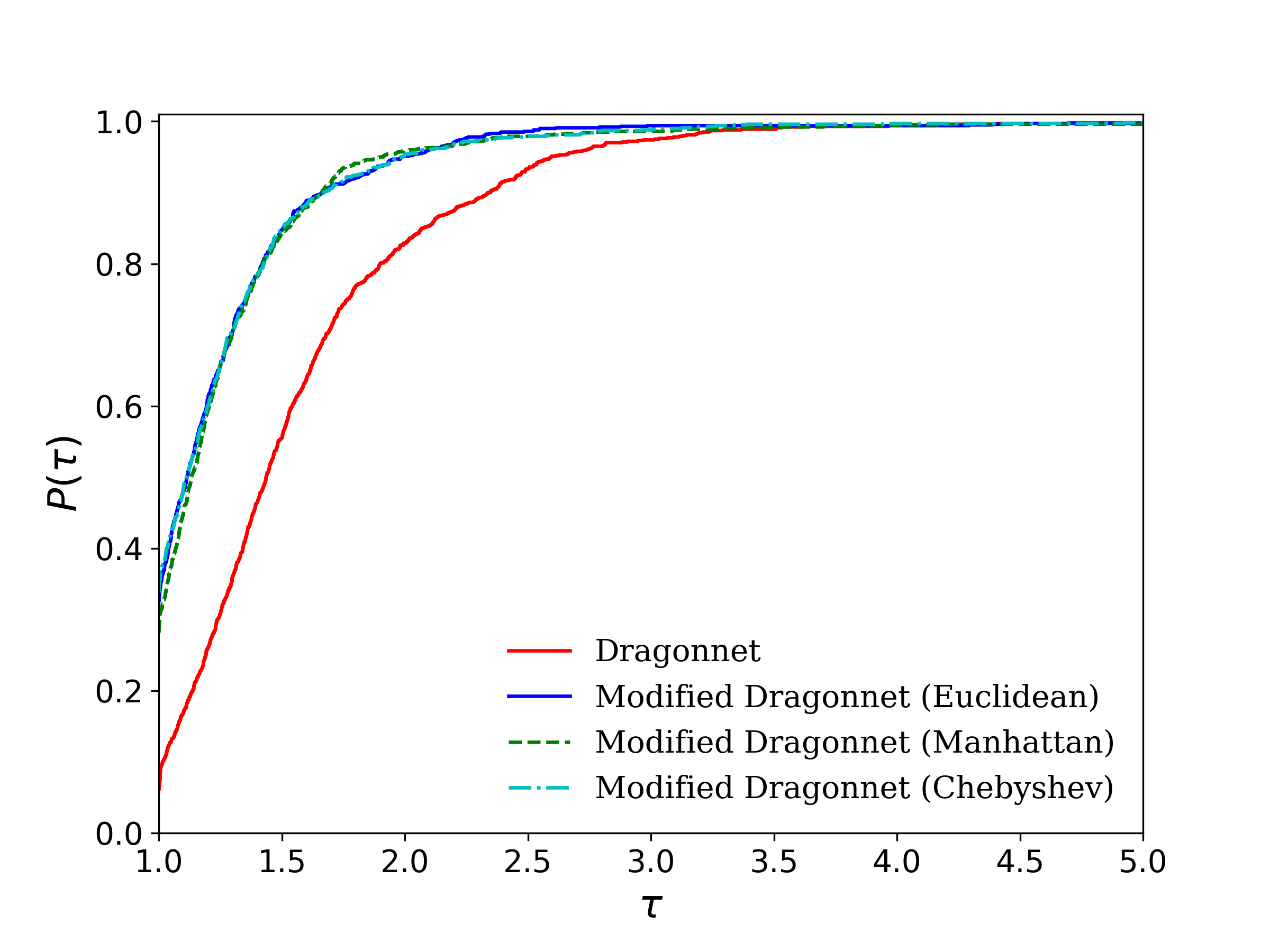}
	\vspace{-.5cm}
	\caption{Performance profiles of all evaluated models based on $\epsilon_{PEHE}$}\label{Fig:Performance profile PEHE}
\end{figure}

Figure \ref{Fig:Performance profile PEHE} presents the performance profiles of the three versions of the proposed model and the\textsf{ Dragonnet}, based on $\epsilon_{PEHE}$ metric. The proposed model considerably outpeformed the state-of-the-art \textsf{Dragonnet} with any used distance metric, in terms of  $\epsilon_{PEHE}$. All versions of \textsf{Modified Dragonnet} solve 34\% of the simulations with the best (lowest) error PEHE, while \textsf{Dragonnet} solves only 8\% of the simulations.

Table \ref{Table:ATE} presents the statistical comparison between the three versions of the proposed model and the \textsf{Dragonnet} based on $|\epsilon_{ATE}|$ metric. Clearly, \textsf{Modified Dragonnet (Euclidean)} reported the best performance, slightly outperforming all compared models. Additionally, it was the only version of the proposed model, which reported better FAR ranking than the state-of-the-art model \textsf{Dragonnet}. However, the interpretation of Finner post-hoc test illustrated that there are not considerable differences, which results that all models performed equally well.

\begin{table}[!ht]
	\centering
	\setlength{\tabcolsep}{10pt}
	\renewcommand{\arraystretch}{1}
	     \begin{tabular}{lccc}
		\toprule
		\multirow{2}{*}{Model}         &  \multirow{2}{*}{FAR} & \multicolumn{2}{c}{Finner post-hoc test}\\
		\cmidrule{3-4}                 &                       & $p_F$-Value &Null hypothesis  \\
		\midrule
		\textsf{Modified Dragonnet (Euclidean)} & 558.768               &      -      &     -           \\
		\textsf{Dragonnet}                      & 561.822               & 0.911658    &  Fail to reject \\
		\textsf{Modified Dragonnet (Manhattan)} & 571.773               & 0.636660    &  Fail to reject \\
		\textsf{Modified Dragonnet (Chebychev)} & 581.637               & 0.406163    &  Fail to reject \\
		\bottomrule
	\end{tabular}
\caption{FAR test and Finner post-hoc test based on $|\epsilon_{ATE}|$}\label{Table:ATE}
\end{table}

Table \ref{Table:PEHE} presents that the proposed model
considerably outpeformed the \textsf{Dragonnet} in terms of $\epsilon_{PEHE}$ with every
utilized distance metric, which is statistically confirmed by FAR and Finner tests. \textsf{Modified Dragonnet (Euclidean)} reported the best performance since it exhibited top ranking. However, Finner post-hoc test reveals that all versions of the model perform equally well and there are no significant statistical differences in their performances.

\begin{table}[!ht]
	\centering
	\setlength{\tabcolsep}{10pt}
	\renewcommand{\arraystretch}{1}
	\begin{tabular}{lccc}
		\toprule
		\multirow{2}{*}{Model}         &  \multirow{2}{*}{FAR} & \multicolumn{2}{c}{Finner post-hoc test}\\
		\cmidrule{3-4}                 &                       & $p_F$-Value &Null hypothesis  \\
		\midrule
		\textsf{Modified Dragonnet (Euclidean)} & 485.004               &      -      &     -           \\
		\textsf{Modified Dragonnet (Chebychev)} & 491.856               &  0.803454   &  Fail to reject \\
		\textsf{Modified Dragonnet (Manhattan)} & 505.099               &  0.465460   &  Fail to reject \\
		\textsf{Dragonnet}                      & 792.042               &      0      &  Reject         \\
		\bottomrule
	\end{tabular}
	\caption{FAR test and Finner post-hoc test based on $\epsilon_{PEHE}$}\label{Table:PEHE}
\end{table}

Based on the previous discussion, we are able to conclude, that the proposed approach estimate PEHE with higher accuracy than state-of-the-art Dragonnet, while it exhibited similar performance regarding the prediction of ATE. This suggests that although the proposed model and Dragonnet report identical performance as estimators, it considerably exhibits better performance as a predictor.

\section{Conclusion}\label{Sec:6}

In this research, we proposed a new neural network model for the prediction of the conditional outcomes and the propensity score as well as the estimation of treatment effects. The architecture of the proposed model constitutes a modification of the state-of-the-art Dragonnet model. An advantage of the proposed model is that it exploits the covariates along with information from the outcomes of the instances contained in the training data.
The motivation of our approach consists of enriching the inputs of the model with the average outcomes of the nearest neighbors from the control and treatment group along with the covariates, in order to improve the prediction performance. 

The experimental analysis demonstrated that the proposed model is a better estimator than Dragonnet, while simultaneously predicts treatment effects with high accuracy. This is confirmed by the performance profiles and the statistical analysis based on a nonparametric and a post-hoc test. It is also worth mentioning that the proposed model exhibited similar performance with the utilization of all three distances.

A limitation of the proposed work is the selection of the optimal value of parameter $k$ and the utilized metric. A study on the efficiency and sensitivity of our approach for different values of parameter $k$ and distance metrics (such as cosine similarity, Jaccard distance and Hamming Distance \cite{pandit2011comparative}) is planned as future work.
Finally, another interesting idea is the adoptation of the proposed approach to other neural network-based models such as TARnet \cite{shalit2017estimating} and NedNet \cite{shi2019adapting} as well as a performance evaluation using other causal modelling benchmarks.

\subsubsection{Acknowledgements} 

The work leading to these results has received funding from the European Union’s Horizon 2020 research and innovation programme under Grant Agreement No. 965231, project REBECCA(REsearch on BrEast Cancer induced chronic conditions supported by Causal Analysis of multi-source data).

\end{document}